\documentclass[letterpaper,10pt,conference]{ieeeconf}

\IEEEoverridecommandlockouts
\usepackage{cite}
\usepackage{graphicx}
\usepackage{amsmath,amssymb,amsfonts}
\usepackage{booktabs}
\usepackage{multirow}
\usepackage{multicol}
\usepackage{adjustbox}
\usepackage{tabularx}
\usepackage[table]{xcolor}
\usepackage{array}
\usepackage{makecell}
\usepackage[caption=false,font=footnotesize]{subfig}
\usepackage{pifont}
\makeatletter
\let\NAT@parse\undefined
\makeatother
\usepackage{hyperref}

\definecolor{rainRefRed}{HTML}{C62828}
\definecolor{rainCiteLavender}{HTML}{9674BA}
\hypersetup{
  colorlinks=true,
  urlcolor=magenta,
  linkcolor=rainRefRed,
  citecolor=rainCiteLavender
}

\newcommand{\figref}[1]{Fig.~\ref{#1}}
\newcommand{\tabref}[1]{Table~\ref{#1}}

\title{RAIN: Robotic Task Generalization with Region-Aware Interaction Networks}
\author{%
Seunghun Lee$^{1,2,*}$, Kyuhwan Shim$^{1,3,*}$, Yeogyeong Song$^{1,*}$, Seung Hwan Kim$^{1}$,\\
Sihaeng Lee$^{1}$, Byoung-Tak Zhang$^{3}$, Sunghoon Im$^{4,\dagger}$, Gunhee Lee$^{1,\dagger}$\\[0.5ex]
{\normalsize $^{1}$LG AI Research \quad $^{2}$Huawei Technologies Ltd. \quad
$^{3}$Seoul National University \quad $^{4}$KAIST}%
\thanks{$^{*}$Equal Contribution. Work done during internships at LG AI Research.}%
\thanks{$^{\dagger}$Corresponding authors.}%
}

\begin{document}

\maketitle
\thispagestyle{plain}
\pagestyle{plain}

\begin{abstract} 
  Existing vision-language-action (VLA) models benefit from large-scale vision-language pretraining, but tend to memorize training trajectories, limiting their ability to adapt learned interactions to new task and scene configurations. To address this problem, we propose \textbf{Region-Aware Interaction Networks (RAIN)}, which formulates a task as a sequence of goal-conditioned subtasks, each specified by an action type and a target-region mask. This formulation allows learned skills to be reused across different objects and scene layouts and composed into task sequences not seen during training. Its Target-adaptive Cross-View Encoder (TCE) conditions multi-view features on target masks and exchanges information across views, while PlanDiT predicts target-directed trajectories and actions. A Transition Head estimates subtask completion to enable automatic transitions. We also introduce a \textbf{Reference-based Retargeting Strategy (RRS)}, which uses recorded trajectories to synthesize smooth approaches to reference interactions with alternative target regions, without collecting additional data. Experiments on \textbf{LIBERO-Analogy}, our new benchmark for robotic task generalization, and real-world tabletop tasks demonstrate that RAIN generalizes more robustly than existing VLA baselines. Project page: \href{https://seung-hun-lee.github.io/projects/RAIN/}{this https URL}.

\end{abstract}
\begin{figure*}[!t]
    \centering
    \includegraphics[width=0.9\linewidth]{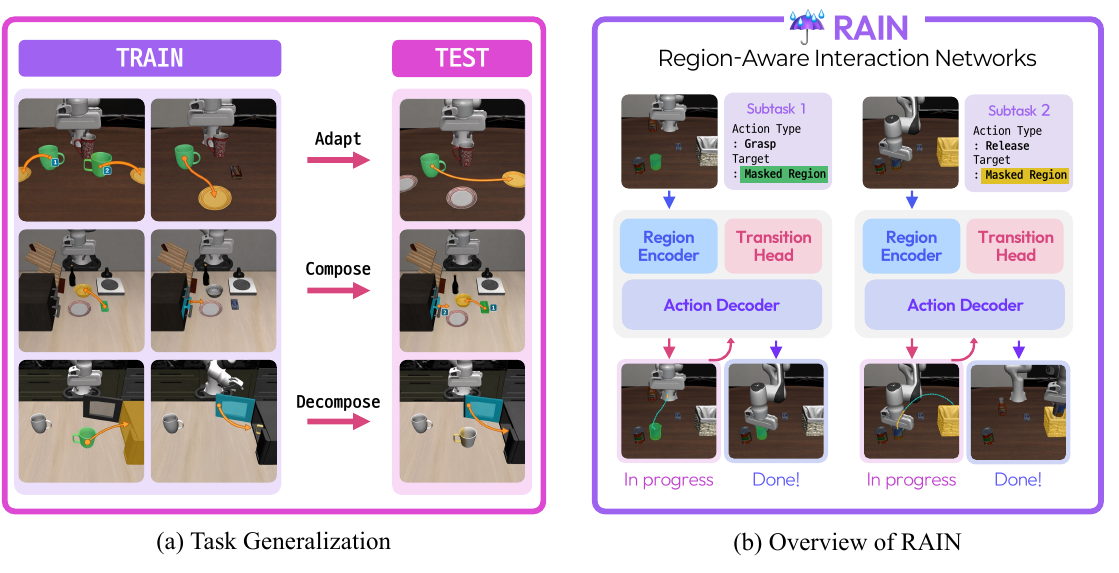}
    \caption{(a) Task generalization by analogy: adapting learned interactions to new task and scene configurations, composing skills, and executing selected parts of learned sequences. (b) \textbf{Region-Aware Interaction Networks (RAIN)} executes each subtask from an action type and target-region mask. A Transition Head determines when to advance to the next subtask.}
    \label{fig:intro}
\end{figure*}

\section{Introduction}
\label{sec:intro}

Robotic manipulation is a central capability for embodied intelligence, with applications spanning industrial automation, household assistance, logistics, and general-purpose service robots. Recent advances in Vision-Language-Models (VLMs) and embodied multimodal models \cite{radford2021learning, liu2023visual, karamcheti2024prismatic, beyer2024paligemma, driess2023palm}, large-scale robot datasets \cite{o2024open, walke2023bridgedata, khazatsky2024droid}, and Vision-Language-Action (VLA) architectures \cite{brohan2022rt, zitkovich2023rt, kim24openvla, octo_2023, black2024pi0, bjorck2025gr00t} have substantially expanded the range of language-conditioned tasks that robot policies can execute from visual observations. By transferring semantic priors from large multimodal pretraining and visuomotor priors from large robot action corpora, these systems have made general robot policies an increasingly active research direction.

The dominant path toward such policies is to couple pretrained visual-language knowledge with action learning, then post-train the resulting system on the target robot tasks \cite{zitkovich2023rt, kim24openvla, kim2025fine, intelligence2025pi_, black2024pi0, zheng2024tracevla, pixelvla2025, huang2025thinkact, kang2024clip}. However, post-training on a narrow set of robot tasks often erodes the general semantic knowledge acquired during pretraining \cite{hancock2025actions}. Even models trained on large-scale vision-language corpora and extensive collections of robot action trajectories, including $\pi_{0.5}$ \cite{intelligence2025pi_}, GR00T N1.6 \cite{bjorck2025gr00t}, and OpenVLA \cite{kim24openvla}, tend to replay memorized trajectories from their training data even when objects are relocated, layouts change, or a new task requires a different use of learned skills \cite{zhou2025libero, fei2025libero}. A general robotic policy should interpret task semantics and infer, by analogy with previously learned tasks, how to apply its skills to tasks outside the training set. We consider three forms of this generalization, as shown in \figref{fig:intro}-(a): (1) \textit{Adapt}, applying learned interactions to new task objectives, object sets, and scene layouts; (2) \textit{Compose}, combining learned skills to solve new long-horizon tasks; and (3) \textit{Decompose}, performing a requested subset of a task learned as a longer sequence. Although recent benchmarks such as LIBERO-Plus \cite{fei2025libero} and LIBERO-PRO \cite{zhou2025libero} examine VLA robustness, their evaluations focus primarily on variations of existing tasks and scenes, leaving broader task generalization by analogy underexplored.

To address this challenge, we propose \textbf{Region-Aware Interaction Networks (RAIN)}, a novel framework for robotic task generalization.
As shown in \figref{fig:intro}-(b), RAIN formulates a task as a sequence of goal-conditioned subtasks, each defined by an action type and a target-region mask.
For example, ``Pick up the Alphabet Soup'' is expressed as ``Action Type: Grasp; Target: Masked Region'' in our formulation.
By grounding each interaction in a visual region rather than a task-specific description, this formulation allows learned skills to be reused across different objects and task sequences.
Our Target-adaptive Cross-View Encoder (TCE) enhances visual features by conditioning each view's features on the target mask via Target-adaptive Layer Normalization (TarLN) and then exchanging information across views to build target-aware representations.
PlanDiT uses these representations to jointly predict a trajectory and an action chunk toward the target.
A Transition Head determines when the current interaction is complete, enabling the policy to advance through the subtask sequence autonomously.
Once trained, RAIN can execute a subtask sequence specified by a human or planned by a VLM from the task description and scene images.
In the latter case, pairing the VLM with SAM~2 \cite{ravi2024sam2} or SAM~3 \cite{carion2025sam3} enables agentic target-region tracking without coupling VLM-based reasoning to action post-training.

Despite this goal-conditioned formulation, training on the same recorded action sequences still risks trajectory memorization.
We therefore introduce a \textbf{Reference-based Retargeting Strategy (RRS)} to learn interactions with alternative target regions without collecting additional data.
RRS uses recorded trajectories involving different objects as references.
At a sampled training timestep, it assigns a target region different from the one in the recorded sequence and synthesizes a smooth approach from the current state into a reference interaction with that target, providing a new action sequence for training.

We evaluate RAIN in both simulation and real-world manipulation. In simulation, we train on 40 LIBERO tasks and evaluate on \textbf{LIBERO-Analogy}, a new benchmark for robotic task generalization. It comprises three suites, Adapt, Compose, and Decompose, with 20 tasks each. RAIN achieves more robust generalization than strong baselines, including robotic foundation models \cite{kim2025fine, intelligence2025pi_, bjorck2025gr00t} and world-action models \cite{kim2026cosmos, sun2026vla, yan2026flex}, while maintaining competitive performance on the original LIBERO suites. Real-world tabletop experiments also evaluate generalization to changes in objects, target locations, and scene layouts.
Our contributions are threefold:
\begin{itemize}
    \item We propose \textbf{RAIN}, a region-aware action model that learns interactions conditioned on an action type and a target-region mask, enabling skill reuse across different objects and task sequences.
    \item We introduce \textbf{RRS}, a reference-based retargeting strategy that synthesizes training trajectories to approach and interact with newly specified target regions.
    \item We introduce \textbf{LIBERO-Analogy}, an evaluation benchmark for robotic task generalization, and demonstrate more robust generalization than strong VLA baselines across its Adapt, Compose, and Decompose suites.
\end{itemize}
\section{Related Works}

\noindent \textbf{Vision-Language-Action Models.}
Vision-language-action (VLA) models~\cite{brohan2022rt,zitkovich2023rt,octo_2023,kim24openvla} transfer pretrained vision-language representations to robot control. Large robot datasets~\cite{o2024open,walke2023bridgedata,khazatsky2024droid} support generalist and flow-based policies across embodiments~\cite{black2024pi0,bjorck2025gr00t,intelligence2025pi_}. Efficient fine-tuning~\cite{kim2025fine} and lightweight architectures~\cite{shukor2025smolvla,lin2026evo} reduce adaptation and deployment costs. Hierarchical and reasoning-augmented methods introduce intermediate plans or reasoning representations~\cite{shi2025hi,li2025hamster,zawalski2024ecot,huang2025thinkact}, while spatially grounded policies incorporate geometry and spatial reasoning into action prediction~\cite{qu2025spatialvla,zhang2025inspire,fang2026molmoact2}. World-action models learn action-relevant dynamics through future visual, geometric, or latent-state prediction~\cite{kim2026cosmos,sun2026vla,yan2026flex}. Recent robustness benchmarks, including LIBERO-PRO~\cite{zhou2025libero} and LIBERO-Plus~\cite{fei2025libero}, test changes in objects, initial states, instructions, layouts, viewpoints, and appearance, but largely preserve the original task semantics. Our LIBERO-Analogy benchmark evaluates adaptation to new task and scene configurations, composition of learned skills into long-horizon tasks, and execution of requested subsets of longer task sequences.

\noindent \textbf{Visual Prompting for Robot Policies.}
Visual and multimodal prompts specify spatial or procedural information for robot policies. VIMA~\cite{jiang2023vima} interleaves language and visual tokens, while subsequent work uses coordinate or sketch-based instructions~\cite{li2024llara,li2025robotic,Tan_2026_CVPR}, trajectory and trace prompts~\cite{zheng2024tracevla}, and dense pixel-level cues such as points, lines, boxes, or masks~\cite{pixelvla2025}. Visual instruction tuning~\cite{liu2023visual} and pixel grounding~\cite{hanoona2023GLaMM} provide foundations for interpreting these prompts.
\begin{figure*}[!t]
    \centering
    \includegraphics[width=0.9\linewidth]{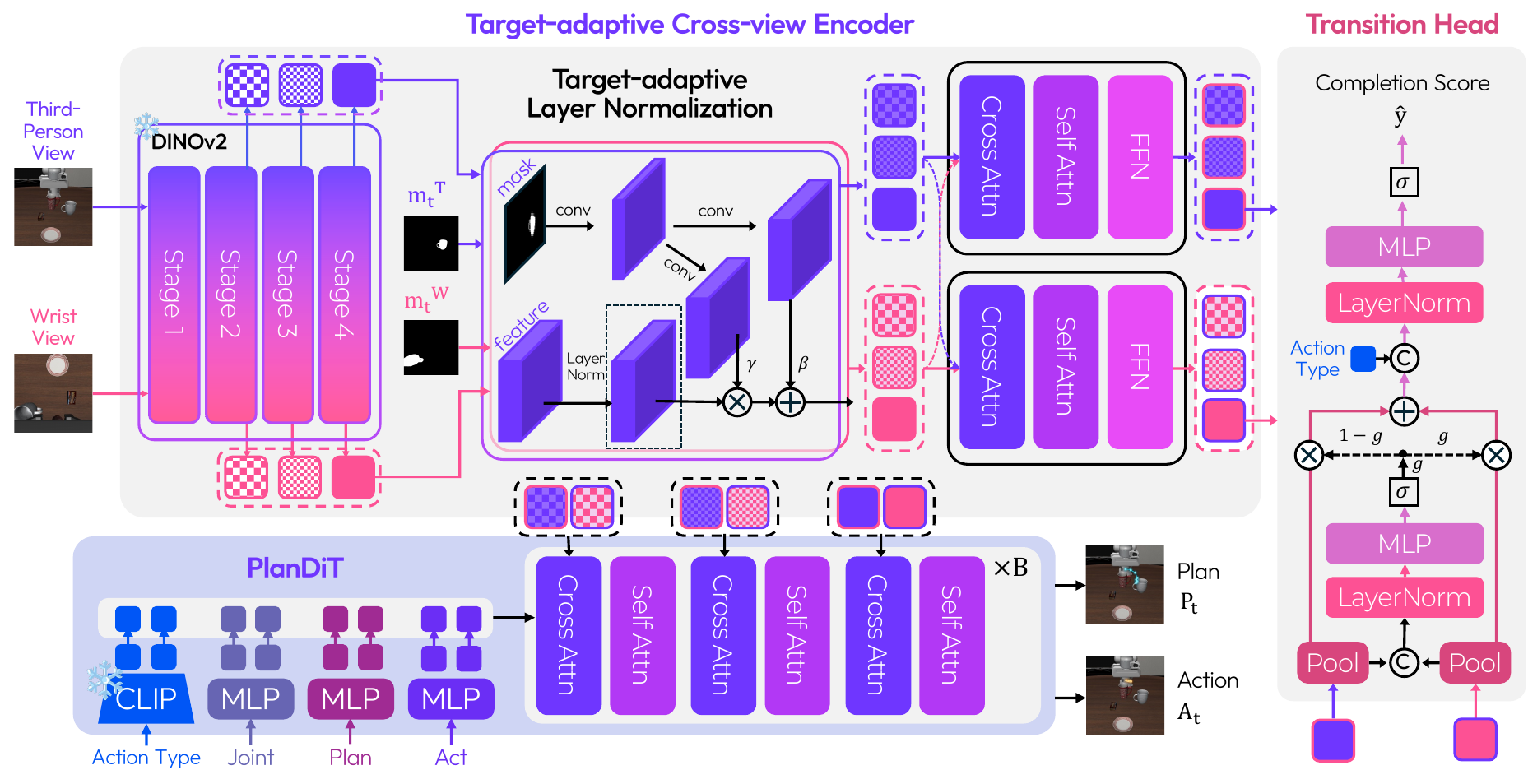}
    \caption{\textbf{Overall architecture of RAIN.} For each subtask, RAIN receives an action type and target masks. The \textbf{Target-adaptive Cross-view Encoder (TCE)} conditions multi-stage features from the third-person and wrist views on the target masks. The \textbf{PlanDiT} uses these stage-wise features to jointly predict future waypoints and action chunks. The \textbf{Transition Head} pools target-region features and fuses the views to predict subtask completion.}
    \label{fig:pipeline}
\end{figure*}

\section{Method}

\subsection{Data Preparation}

Qwen~\cite{yang2025qwen3} decomposes each task into target-centered subtasks, forming an ordered sequence of \textit{(action type, target)} pairs. We determine subtask intervals from recorded gripper trajectories: finger closing, opening, and stabilization identify grasp and release phases, while changes or stops in translation and rotation delimit other interactions. The detected completion event defines the interval endpoint, and this final frame receives a \textit{completion} label. For simulated demonstrations, we obtain target masks in each view from the simulator's segmentation output. For real-world demonstrations, SAM~2~\cite{ravi2024sam2} segments and tracks the targets to produce mask trajectories.

\subsection{Region-Aware Interaction Networks}

\figref{fig:pipeline} illustrates the overall architecture of RAIN. At each timestep $t$, RAIN observes multi-view RGB images $\mathcal{I}_t$ and corresponding target masks $\mathcal{M}_t$, together with the proprioceptive state $\mathbf{q}_t\in\mathbb{R}^{8}$ and an action type $c_t$. For clarity, the equations below use third-person ($T$) and wrist ($W$) views, but additional cameras can be included, as in our three-view real-world setup. The state contains the end-effector position, its axis-angle orientation, and the two gripper-finger positions. The model aims to learn $p(\mathbf{A}_t,\mathbf{P}_t \mid \mathcal{I}_t,\mathbf{q}_t,c_t,\mathcal{M}_t)$, where $\mathbf{A}_t=[\mathbf{a}_t,\ldots,\mathbf{a}_{t+H-1}]\in\mathbb{R}^{H\times7}$ is an action chunk of horizon $H$ and $\mathbf{P}_t=\{\mathbf{p}_{t,1},\ldots,\mathbf{p}_{t,N}\}\in\mathbb{R}^{N\times3}$ is an auxiliary plan with $N$ waypoints. Each action contains relative 3D translation and rotation commands and a gripper command. The plan represents target-directed 3D waypoints and guides the action decoder toward the region specified by the masks.

\noindent\textbf{Target-adaptive Cross-view Encoder.}
TCE extracts features $\mathbf{f}^v_{t,\ell}\in\mathbb{R}^{h\times w\times d}$ from the last three stages of a frozen DINOv2 encoder~\cite{oquab2023dinov2}, where $v\in\{T,W\}$ and $\ell\in\{2,3,4\}$. After resizing the target masks to the feature resolution, Target-adaptive Layer Normalization (TarLN) conditions each stage on its view's mask:
\begin{equation}
    \bar{\mathbf{f}}^v_{t,\ell}
    =\gamma^v(\mathbf{m}^v_t)\odot\mathrm{LN}(\mathbf{f}^v_{t,\ell})
      +\beta^v(\mathbf{m}^v_t).
\end{equation}
Here, $\mathrm{LN}$ denotes layer normalization~\cite{ba2016layer}, and $\gamma^v$ and $\beta^v$ map the mask to spatial scaling and bias, following spatially-adaptive normalization~\cite{park2019semantic}.

At each stage, a cross-view Transformer block applies cross-attention between the two views, followed by self-attention and a feed-forward network. Cross-attention uses the first argument as the query and the second as keys and values:
\begin{equation}
\begin{aligned}
\tilde{\mathbf{f}}^T_{t,\ell}
    &=\mathrm{FFN}(\mathrm{SelfAttn}(\mathrm{CrossAttn}(
        \bar{\mathbf{f}}^T_{t,\ell},\bar{\mathbf{f}}^W_{t,\ell}))), \\
\tilde{\mathbf{f}}^W_{t,\ell}
    &=\mathrm{FFN}(\mathrm{SelfAttn}(\mathrm{CrossAttn}(
        \bar{\mathbf{f}}^W_{t,\ell},\bar{\mathbf{f}}^T_{t,\ell}))).
\end{aligned}
\end{equation}
TCE parameters are shared across stages. We concatenate the two views at each stage as $\phi^\ell_t=[\tilde{\mathbf{f}}^T_{t,\ell};\tilde{\mathbf{f}}^W_{t,\ell}]$, yielding the context $\Phi_t=\{\phi^2_t,\phi^3_t,\phi^4_t\}$.

\noindent\textbf{Plan-guided Diffusion Transformer.}
PlanDiT is a generative Transformer that predicts a target-directed geometric plan and an executable action chunk. The plan $\mathbf{P}_t$ consists of 3D waypoints toward the target region, with the final waypoint $\mathbf{p}_{t,N}$ specifying the interaction point. 

PlanDiT forms an input token sequence $\mathbf{E}_t=[\mathbf{e}_{c};\mathbf{e}_{q};\mathbf{e}_{p};\mathbf{e}_{a}]$, where $\mathbf{e}_{c}$ is the action-type feature from frozen CLIP~\cite{radford2021learning}, $\mathbf{e}_{q}$ is the proprioception embedding, and $\mathbf{e}_{p},\mathbf{e}_{a}$ are plan and action tokens embedded from the noisy sample used in flow matching. Input projections are omitted from the notation. The decoder stacks $B$ cycles, each comprising three cross-attention/self-attention pairs. Within each cycle, the cross-attention layers use the projected TCE contexts $\phi^2_t,\phi^3_t,\phi^4_t$ as keys and values in that order. Self-attention uses an asymmetric mask: action tokens can attend to plan tokens, but plan tokens cannot attend to action tokens.

We train PlanDiT with Flow Matching \cite{bjorck2025gr00t}. Let $\mathbf{x}_t=[\mathbf{P}_t;\mathbf{A}_t]$ denote the joint plan and action target. Given a flow step $\tau\in[0,1]$ and Gaussian noise $\epsilon\sim\mathcal{N}(\mathbf{0},\mathbf{I})$, we define $\mathbf{x}^{\tau}_t=\tau\mathbf{x}_t+(1-\tau)\epsilon$ and optimize the vector field $V_{\theta}$ by
\begin{equation}
\begin{aligned}
    \mathcal{L}_{\mathrm{fm}}=\mathbb{E}_{\tau,\epsilon,\mathbf{x}_t}\big[
    \big\|V_{\theta}(\mathbf{x}^{\tau}_t,\tau,\Phi_t,\mathbf{q}_t,c_t)
    -(\mathbf{x}_t-\epsilon)\big\|_2^2\big].
\end{aligned}
\end{equation}
We further enforce consistency of the final waypoint across frames from the same subtask. We compare predicted endpoints $\hat{\mathbf{p}}_{t,N}$ in a common coordinate frame. For $K$ sampled frames with $t_K$ selected as the reference closest to the demonstrated interaction position, we minimize
\begin{equation}
    \mathcal{L}_{\mathrm{cons}}=\frac{1}{K-1}\sum_{k=1}^{K-1}
    \mathrm{SmoothL1}(\hat{\mathbf{p}}_{t_k,N},\hat{\mathbf{p}}_{t_K,N}).
\end{equation}
The total objective for PlanDiT is $\mathcal{L}_{\mathrm{PlanDiT}}=\lambda_{\mathrm{fm}}\mathcal{L}_{\mathrm{fm}}+\lambda_{\mathrm{cons}}\mathcal{L}_{\mathrm{cons}}$, where $\lambda_{\mathrm{fm}}$ and $\lambda_{\mathrm{cons}}$ are loss weights.

\noindent\textbf{Transition Head.}
Since TCE has already enhanced the target-region features with information across views, the Transition Head pools these features from the final stage:
\begin{equation}
    \mathbf{z}^v_t=\mathrm{Proj}\!\left(
    \frac{\mathrm{Avg}(\mathbf{m}^v_t\odot\tilde{\mathbf{f}}^v_{t,4})}
         {\mathrm{Avg}(\mathbf{m}^v_t)}\right),
\end{equation}
where $\mathrm{Avg}$ averages over spatial locations and $\mathrm{Proj}$ is a shared linear projection. The denominator normalizes by the target area. An empty mask uses global average pooling.

The views can contribute differently to subtask completion judgments. We therefore learn a scalar gate to fuse their pooled features and concatenate the action-type feature $\mathbf{e}_c$ defined above:
\begin{equation}
\begin{aligned}
    g_t&=\sigma(\mathrm{MLP}(\mathrm{LN}([\mathbf{z}^T_t;\mathbf{z}^W_t]))),\\
    \mathbf{z}_t&=(1-g_t)\mathbf{z}^T_t+g_t\mathbf{z}^W_t,\\
    \hat{y}_t&=\sigma(\mathrm{MLP}(\mathrm{LN}([
        \mathbf{z}_t;\mathbf{e}_c]))).
\end{aligned}
\end{equation}
Here, $\sigma$ is the sigmoid function.

After action training, we freeze TCE and PlanDiT and train the Transition Head with weighted binary cross entropy:
\begin{equation}
    \mathcal{L}_{\mathrm{head}}=
    \mathbb{E}_t\big[w_t\,\mathrm{BCE}(\hat{y}_t,y_t)\big],
\end{equation}
where $y_t$ is the completion label and $w_t$ upweights selected completed samples. During rollout, RAIN advances to the next subtask when the completion probability exceeds a threshold for consecutive predictions.

\subsection{Reference-based Retargeting Strategy}

RRS expands region-aware interaction coverage by generating trajectories toward target configurations that are not directly covered by the original demonstrations. Given a recorded reference trajectory $\mathcal{R}=\{\mathbf{r}_0,\ldots,\mathbf{r}_{M-1}\}$ and an arbitrary approach state $\mathbf{x}_0\in\mathbb{R}^{3}$ denoting an end-effector position, RRS first identifies a tangent junction $\mathbf{r}_{k^*}$ on the reference trajectory. For each reference point $k\in\{1,\ldots,M-2\}$, we compute the unit tangent vector $\mathbf{t}_k=(\mathbf{r}_{k+1}-\mathbf{r}_{k-1})/\|\mathbf{r}_{k+1}-\mathbf{r}_{k-1}\|_2$, the approach direction $\mathbf{v}_k=(\mathbf{r}_k-\mathbf{x}_0)/\|\mathbf{r}_k-\mathbf{x}_0\|_2$, and the distance $d_k=\|\mathbf{r}_k-\mathbf{x}_0\|_2$. The junction score is
\begin{equation}
    S_k=(\mathbf{v}_k^\top\mathbf{t}_k)\left(1-\frac{d_k}{d_{\max}}\right)\left(1-\frac{k}{M}\right),
\end{equation}
and the junction index is selected as $k^*=\operatorname*{arg\,max}_{k} S_k$ under proximity and directional-alignment constraints, where $d_{\max}$ is a normalization radius determined by the workspace scale. The first term favors tangential merging into the reference motion, the second avoids long approach paths, and the third preserves as much of the successful reference interaction as possible.

Once the junction is chosen, RRS connects $\mathbf{x}_0$ to $\mathbf{r}_{k^*}$ using a cubic Hermite spline \cite{kochanek1984interpolating}. The variable $u\in[0,1]$ denotes the normalized interpolation parameter:
\begin{equation}
\begin{aligned}
    \mathbf{C}(u)
        &=\psi_{00}(u)\mathbf{x}_0+\psi_{10}(u)\mathbf{D}_0 \\
        &\quad+\psi_{01}(u)\mathbf{r}_{k^*}+\psi_{11}(u)\mathbf{D}_1,
        \quad u\in[0,1],
\end{aligned}
\end{equation}
where $\mathbf{D}_0=\mathbf{r}_{k^*}-\mathbf{x}_0$, $\mathbf{D}_1=L\mathbf{t}_{k^*}$, and $L=\|\mathbf{r}_{k^*}-\mathbf{x}_0\|_2$. The basis functions are defined as follows:
\begin{equation}
\begin{aligned}
    \psi_{00}&=2u^3-3u^2+1, &
    \psi_{10}&=u^3-2u^2+u, \\
    \psi_{01}&=-2u^3+3u^2, &
    \psi_{11}&=u^3-u^2.
\end{aligned}
\end{equation}
The approach segment is discretized with $N_{\mathrm{app}}$ steps using $u_i=i/N_{\mathrm{app}}$ for $i=0,\ldots,N_{\mathrm{app}}$. The final trajectory is obtained by concatenating the generated approach and the remaining reference interaction:
\begin{equation}
    \mathcal{P}=[\mathbf{C}(u_0),\ldots,\mathbf{C}(u_{N_{\mathrm{app}}}),\mathbf{r}_{k^*+1},\ldots,\mathbf{r}_{M-1}].
\end{equation}
We convert the approach segment into delta-position actions and interpolate orientation toward the junction pose. After the junction, the recorded reference actions are reused, preserving contact-rich interaction dynamics. At the current observation, we condition RAIN on the alternative target region and use the synthesized action trajectory as supervision.

\begin{table*}[t]
\centering
\caption{\textbf{Comparison on LIBERO and LIBERO-Analogy.} 
We report success rates on the original LIBERO suites and our LIBERO-Analogy 
task-generalization benchmark. $^{\dagger}$ denotes our reproduction using 
a single joint 4-suite fine-tuned model.}
\label{tab:supp_libero}

\setlength{\tabcolsep}{2.8pt}
\renewcommand{\arraystretch}{1.1}

\resizebox{0.99\linewidth}{!}{%
\begin{tabular}{
l
c c
c c c c
c c c c c
}
\toprule
& \multicolumn{2}{c}{\textbf{Trainable Params}}
& \multicolumn{4}{c}{\textbf{LIBERO-Analogy}}
& \multicolumn{5}{c}{\textbf{LIBERO}} \\
\cmidrule(lr){2-3}
\cmidrule(lr){4-7}
\cmidrule(lr){8-12}

\multirow{-2}{*}{\textbf{Method}}
& \textbf{Pre-train}
& \textbf{Post-train}
& \textbf{Adapt}
& \textbf{Compose}
& \textbf{Decompose}
& \textbf{Avg.}
& \textbf{Spatial}
& \textbf{Object}
& \textbf{Goal}
& \textbf{Long}
& \textbf{Avg.} \\
\midrule

\multicolumn{12}{l}{\textbf{\textit{Regular VLA Models}}} \\[1pt]

SmolVLA \cite{shukor2025smolvla}
{\color{gray}\scriptsize [arXiv'25]}
& 0.1B & 0.1B
& 5.6 & 0.0 & 5.4 & 3.7
& 90.0 & 96.0 & 92.0 & 71.0 & 87.3 \\

OpenVLA-OFT \cite{kim2025fine}
{\color{gray}\scriptsize [RSS'25]}
& 7B & 0.3B
& 18.4 & 6.0 & 6.4 & 10.3
& 97.6 & 98.4 & 97.9 & 94.5 & 97.1 \\

GR00T N1.6$^{\dagger}$ \cite{bjorck2025gr00t}
{\color{gray}\scriptsize [arXiv'25]}
& 3B & 3B
& 27.5 & 10.6 & 9.8 & 16.0
& 96.6 & \textbf{100.0} & 95.8 & 93.0 & 96.3 \\

$\pi_{0}$ \cite{black2024pi0}
{\color{gray}\scriptsize [RSS'25]}
& 3B & 3B
& 13.6 & 1.1 & 15.5 & 10.1
& 96.8 & 98.8 & 95.8 & 85.2 & 94.2 \\

$\pi_{0.5}$ \cite{intelligence2025pi_}
{\color{gray}\scriptsize [CoRL'25]}
& 3B & 3B
& \underline{57.1} & 10.7 & 7.6 & 25.1
& \underline{98.8} & 98.2 & 98.0 & 92.4 & 96.9 \\

Evo-1 \cite{lin2026evo}
{\color{gray}\scriptsize [CVPR'26]}
& 0.1B & 0.8B
& 24.0 & 0.3 & 9.5 & 11.3
& 92.7 & 97.7 & 96.3 & 92.3 & 94.8 \\

\midrule[1pt]

\multicolumn{12}{l}{\textbf{\textit{World-Action Models}}} \\[1pt]

Cosmos Policy \cite{kim2026cosmos}
{\color{gray}\scriptsize [ICLR'26]}
& \textsc{-} & 2B
& 29.1 & 5.0 & 7.1 & 13.7
& 98.1 & \textbf{100.0} & \underline{98.2} & \underline{97.6} & \underline{98.5} \\

VLA-JEPA \cite{sun2026vla}
{\color{gray}\scriptsize [ECCV'26]}
& 2.4B & 2.4B
& 42.1 & 5.8 & 7.8 & 18.6
& 96.2 & 99.6 & 97.2 & 95.8 & 97.2 \\

Flex-$\pi$ \cite{yan2026flex}
{\color{gray}\scriptsize [arXiv'26]}
& 6B & 6B
& 36.7 & 11.3 & 11.1 & 19.7
& \textbf{99.2} & \underline{99.8} & \textbf{99.0} & \textbf{98.6} & \textbf{99.2} \\

\midrule[1pt]

\multicolumn{12}{l}{\textbf{\textit{VLA Models with Spatial Reasoning}}} \\[1pt]

InspireVLA \cite{zhang2025inspire}
{\color{gray}\scriptsize [ICRA'26]}
& 1B & 1B
& 29.2 & 0.0 & 66.4 & 31.9
& 90.7 & 94.3 & 88.3 & 73.3 & 86.7 \\

Action-Sketcher \cite{Tan_2026_CVPR}
{\color{gray}\scriptsize [CVPR'26]}
& 3B & 3B
& 35.8 & 2.0 & 6.6 & 14.8
& 97.2 & 99.6 & 94.8 & 96.0 & 96.9 \\

MolmoAct2 \cite{fang2026molmoact2}
{\color{gray}\scriptsize [CoRL'26]}
& 4.4B & 5B
& 53.9 & 29.5 & 22.4 & 35.3
& 97.8 & \textbf{100.0} & 97.8 & 93.2 & 97.2 \\

\rowcolor[gray]{0.95}
\textbf{\textsc{RAIN}} {\scriptsize (Qwen3.5-4B~\cite{qwen3.5} + SAM3~\cite{carion2025sam3})}
& - & 0.2B
& 53.3 & \underline{31.0} & \underline{77.5} & \underline{53.9}
& 92.4 & 96.8 & 91.2 & 82.4 & 90.7 \\

\rowcolor[gray]{0.95}
\textbf{\textsc{RAIN} \scriptsize{(with GT masks)}}
& - & 0.2B
& \textbf{62.4} & \textbf{37.1} & \textbf{82.7} & \textbf{60.7}
& 93.6 & 99.0 & 95.2 & 93.6 & 95.4 \\

\bottomrule
\end{tabular}
}
\end{table*}
\begin{table*}[t]
\centering
\caption{\textbf{VLM comparison on LIBERO and LIBERO-Analogy.}
All VLMs use SAM3~\cite{carion2025sam3} for segmentation and tracking.
SR (\%) and tracking $\mathcal{J}\&\mathcal{F}$ (\%) use the same rollouts. $\mathcal{J}\&\mathcal{F}$ pools visible-target
observations across both views and tasks. }
\label{tab:vlm_sr_jf}
\scriptsize
\setlength{\tabcolsep}{2pt}
\renewcommand{\arraystretch}{1.16}
\begin{tabular*}{\linewidth}{@{\extracolsep{\fill}}l*{18}{c}@{}}
\toprule
\multirow{3}{*}{\textbf{Method}}
& \multicolumn{10}{c}{\textbf{LIBERO}}
& \multicolumn{8}{c}{\textbf{LIBERO-Analogy}} \\
\cmidrule(lr){2-11}\cmidrule(lr){12-19}
& \multicolumn{2}{c}{Spatial}
& \multicolumn{2}{c}{Object}
& \multicolumn{2}{c}{Goal}
& \multicolumn{2}{c}{Long}
& \multicolumn{2}{c}{Overall}
& \multicolumn{2}{c}{Adapt}
& \multicolumn{2}{c}{Compose}
& \multicolumn{2}{c}{Decompose}
& \multicolumn{2}{c}{Overall} \\
\cmidrule(lr){2-3}\cmidrule(lr){4-5}\cmidrule(lr){6-7}
\cmidrule(lr){8-9}\cmidrule(lr){10-11}\cmidrule(lr){12-13}
\cmidrule(lr){14-15}\cmidrule(lr){16-17}\cmidrule(lr){18-19}
& SR & $\mathcal{J}\&\mathcal{F}$ & SR & $\mathcal{J}\&\mathcal{F}$ & SR & $\mathcal{J}\&\mathcal{F}$ & SR & $\mathcal{J}\&\mathcal{F}$ & SR & $\mathcal{J}\&\mathcal{F}$
& SR & $\mathcal{J}\&\mathcal{F}$ & SR & $\mathcal{J}\&\mathcal{F}$ & SR & $\mathcal{J}\&\mathcal{F}$ & SR & $\mathcal{J}\&\mathcal{F}$ \\
\midrule
Qwen3.5-4B \cite{qwen3.5}    & 92.4 & 80.0 & 96.8 & 76.1 & 91.2 & 55.9 & 82.4 & 57.8 & 90.7 & 65.4 & 53.3 & 60.9 & 31.0 & 51.9 & 77.5 & 49.4 & 53.9 & 53.7 \\
Qwen3.5-9B \cite{qwen3.5}    & 91.6 & 80.7 & 95.8 & 75.4 & 90.0 & 55.3 & 81.4 & 57.6 & 89.7 & 65.2 & 53.2 & 59.5 & 30.0 & 52.8 & 77.4 & 49.8 & 53.5 & 54.0 \\
Cosmos-Reason2-8B \cite{nvidia2026cosmosreason2}  & 87.6 & 83.3 & 90.2 & 72.2 & 87.4 & 55.7 & 83.0 & 59.8 & 87.1 & 66.3 & 60.6 & 63.4 & 29.9 & 56.4 & 76.2 & 51.8 & 55.6 & 57.3 \\
\bottomrule
\end{tabular*}
\end{table*}

\section{Experiments}

\subsection{Implementation Details}

\noindent\textbf{General.}
We use frozen DINOv2-L/14~\cite{oquab2023dinov2} for visual encoding with feature dimension $d=1024$ and frozen CLIP ViT-L/14~\cite{radford2021learning} for action-type encoding. PlanDiT uses $B=2$ stage cycles, giving six cross-attention and six self-attention layers, with action horizon $H=16$ and $N=8$ plan waypoints. The consistency loss uses $K=4$ frames, with $\lambda_{\mathrm{fm}}=1$ and $\lambda_{\mathrm{cons}}=0.1$. Inference uses four flow-matching steps and replans every eight actions. The default transition criterion is a completion probability above 0.7 for two consecutive predictions. After action training, we train the Transition Head for up to 30 epochs (about 7.5k iterations in simulation). For Transition Head training, we hold out 10\% of the data for validation, select the checkpoint with the highest validation F1 score, and stop early if this score does not improve for eight consecutive epochs. Transition Head training uses sample weights of 1.5 for selected completed samples and 1 otherwise. Both training stages use AdamW~\cite{loshchilov2017decoupled} with weight decay 0.01, a 2,000-iteration warmup to $10^{-4}$, and cosine learning-rate decay to $10^{-6}$. RRS is applied only to feasible combinations of the gripper state and target; for example, placement retargeting requires a held object and an alternative valid destination. Inspired by \cite{jin2026agentrvos}, our pipeline first uses Qwen3.5~\cite{qwen3.5} to plan subtasks. SAM~3~\cite{carion2025sam3} then proposes mask candidates for each target, and Qwen selects the most appropriate candidate. SAM~3 tracks the selected mask online as the rollout proceeds.

\noindent\textbf{Simulation.}
RGB images are resized to $224\times224$, giving TCE features with $h=w=16$. Both training stages use a batch size of 1024. Action training runs for 100k iterations.

\noindent\textbf{Real-world.}
We train RAIN for 30k iterations with three RGB views resized to $448\times448$, a total batch size of 64.

\subsection{Simulation Experiments}

\noindent\textbf{Task.}
We train a single RAIN policy on the 40 tasks from the LIBERO Spatial, Object, Goal, and Long suites. LIBERO-Analogy evaluates generalization to new tasks through three suites, Adapt, Compose, and Decompose, with 20 tasks each. \textbf{We strongly recommend watching our supplementary video to see the individual task configurations.} We evaluate RAIN over 50 episodes per task and report success rate (SR). For VLM-based masks, we also measure tracking $\mathcal{J}\&\mathcal{F}$ on the same rollouts. GT and predicted-mask comparisons use the same RAIN checkpoint and initial states.

\noindent\textbf{Results.}
\tabref{tab:supp_libero} compares RAIN with standard VLA models, world-action models (WAMs), and models with spatial reasoning. Among standard VLAs, $\pi_{0.5}$~\cite{intelligence2025pi_}, whose training includes semantic subtask prediction, achieves the highest average SR on LIBERO-Analogy. However, on Compose and Decompose, it tends to replay memorized training trajectories even when the instruction changes the task semantics. Among WAMs, Flex-$\pi$~\cite{yan2026flex} incorporates 3D geometry into future prediction and offers the strongest overall balance across suites, but its average SR remains low at 19.7\%.

Among spatial-reasoning models, InspireVLA~\cite{zhang2025inspire} learns intrinsic spatial reasoning and is particularly effective at executing the requested subset of skills on Decompose, reaching 66.4\% SR, substantially above the other baselines. MolmoAct2~\cite{fang2026molmoact2}, a recent state-of-the-art action-reasoning model, achieves strong results on Adapt and Compose. Its large-scale training in multimodal reasoning and geometric perception likely contributes to these results, although its Decompose SR remains low at 22.4\%.

With GT masks, RAIN achieves the highest SR in all three LIBERO-Analogy suites, averaging 60.7\%. Using Qwen3.5-4B~\cite{qwen3.5} and SAM~3~\cite{carion2025sam3} for task planning and goal-mask tracking, RAIN retains an average SR of 53.9\%, outperforming all baselines overall. On LIBERO, the corresponding SRs are 95.4\% with GT masks and 90.7\% with predicted masks.

\noindent\textbf{VLM Variants.}
\tabref{tab:vlm_sr_jf} reports target-region tracking $\mathcal{J}\&\mathcal{F}$ alongside SR for each suite on the same simulation rollouts. We compare predicted masks with simulator-provided GT masks at every policy observation and aggregate scores over visible targets in both views; GT masks are used only for scoring. The largest drop from GT masks occurs on LIBERO Long, from 93.6\% to 82.4\% SR with Qwen3.5-4B. Long-horizon success requires every subtask to succeed, while maintaining a referred target through occlusion and disappearance remains difficult in long videos~\cite{liang2026long}. Longer rollouts therefore provide more opportunities for tracking errors to interrupt task execution.

\begin{table}[!t]
\centering
\caption{\textbf{Ablation studies.} SR (\%) on LIBERO and LIBERO-Analogy.}
\label{tab:ablation_all_in_one_expanded}
\footnotesize
\setlength{\tabcolsep}{2.5pt}
\renewcommand{\arraystretch}{1.04}

\textbf{(a) Main components}\par\smallskip
\begin{tabular*}{\linewidth}{@{\extracolsep{\fill}}lcc@{}}
\toprule
\textbf{Variant} & \textbf{LIBERO} & \textbf{LIBERO-Analogy} \\
\midrule
Full & \textbf{95.4} & \textbf{60.7} \\
\midrule
\multicolumn{3}{l}{\textbf{TCE}} \\
w/o cross-view & \underline{94.4} & 50.2 \\
w/o multi-stage features & 89.2 & 35.0 \\
\midrule
\multicolumn{3}{l}{\textbf{PlanDiT}} \\
w/o plan token & 91.8 & 40.8 \\
w/o $\mathcal{L}_{\mathrm{cons}}$ & 93.9 & 51.6 \\
w/ bi-directional attention & 94.3 & \underline{51.9} \\
\midrule
\multicolumn{3}{l}{\textbf{Retarget strategy}} \\
w/o retarget & 92.5 & 32.6 \\
Linear interpolation & 91.4 & 41.3 \\
\bottomrule
\end{tabular*}

\par\medskip
\textbf{(b) Transition head}\par\smallskip
\begin{tabularx}{\linewidth}{Xlcc}
\toprule
\textbf{Pooling} & \textbf{Fusion} & \textbf{LIBERO} & \textbf{LIBERO-Analogy} \\
\midrule
Global & Concat & 94.6 & 55.7 \\
Global & Gated & 94.4 & 53.2 \\
Target region & Concat & \underline{95.2} & 59.7 \\
\rowcolor[gray]{0.95}
Target region & Gated & \textbf{95.4} & \textbf{60.7} \\
Global + target region & Concat & 95.0 & \underline{60.1} \\
Global + target region & Gated & \underline{95.2} & 57.7 \\
\bottomrule
\end{tabularx}
\end{table}

\noindent\textbf{Ablation Study.}
All ablations use GT target masks to assess the performance ceiling of each model variant under ideal target localization, isolating component effects from mask-prediction errors.

\tabref{tab:ablation_all_in_one_expanded}-(a) shows effectiveness of our main components. Removing multi-stage features reduces SR from 95.4\% to 89.2\% on LIBERO and from 60.7\% to 35.0\% on LIBERO-Analogy. Since LIBERO-Analogy applies interactions learned on LIBERO to new tasks, weaknesses in these basic interactions also limit generalization. Cross-view processing further improves performance on both benchmarks.

Plan tokens provide goal awareness by representing a target-directed trajectory for interaction with the specified region; removing them reduces LIBERO-Analogy SR to 40.8\%. The asymmetric attention mask encodes a natural goal-to-action dependency: action tokens attend to the plan, while plan tokens do not attend to the actions. Bidirectional attention blurs this separation and lowers SR to 51.9\%. Endpoint consistency also improves generalization by encouraging a stable interaction goal across observations.

Even without retargeting, the goal-conditioned RAIN formulation achieves 32.6\% SR on LIBERO-Analogy, exceeding most baselines in \tabref{tab:supp_libero}. Simple linear retargeting, which generates a straight trajectory from the current gripper position to an alternative interaction goal, raises this to 41.3\%. RRS further improves SR to 60.7\% by smoothly connecting the current state to recorded interaction trajectories, and outperforms linear retargeting on both benchmarks.

For the Transition Head, \tabref{tab:ablation_all_in_one_expanded}-(b) shows that target-region pooling outperforms global pooling with either fusion rule. Global pooling mixes the target with unrelated content, making the completion representation less target-specific. Target-region pooling with gated fusion performs best. With the same gated fusion, concatenating global and target-region pooled features reduces LIBERO-Analogy SR from 60.7\% to 57.7\%, suggesting that the additional context dilutes the target-focused representation already established by TCE.

\begin{figure*}[!t]
    \centering
    \includegraphics[width=0.9\linewidth]{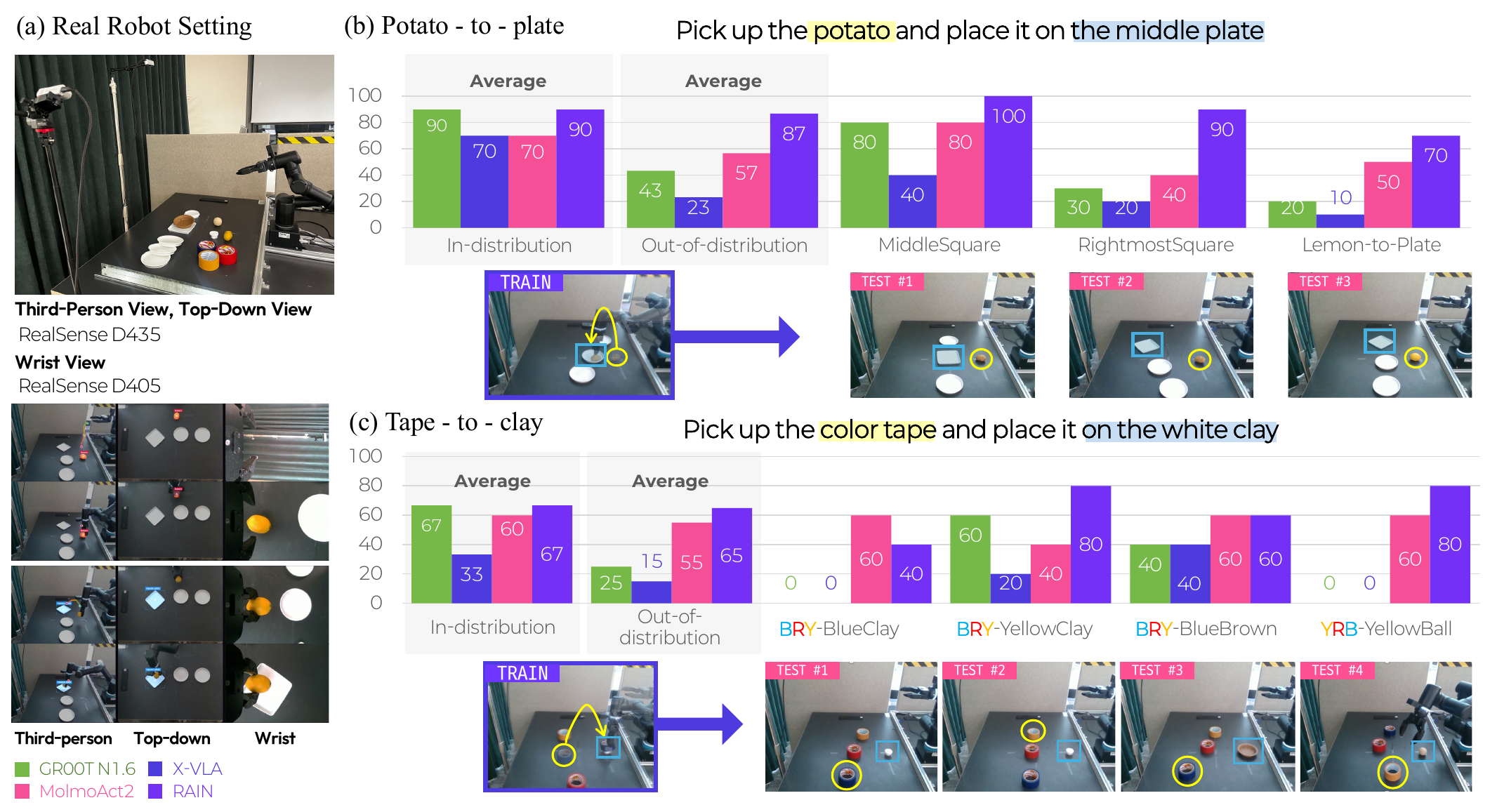}
    \caption{\textbf{Real-world setup and quantitative results.} \textbf{(a)} Robot and camera setup. Success rates for \textbf{(b)} \texttt{Potato-to-Plate} and \textbf{(c)} \texttt{Tape-to-Clay} under in-distribution and out-of-distribution conditions.}
    \label{fig:exp_real_world}
    \vspace{-3mm}
\end{figure*}

\subsection{Real-World Experiments}

\noindent\textbf{Setting.}
As shown in \figref{fig:exp_real_world}-(a), our platform uses a 6-DoF ROBOTIS OMY-F3M arm with a two-finger gripper, a 7D joint-space action interface, and RGB observations from wrist-mounted, third-person, and top-down cameras. Demonstrations are recorded through ROS and converted to LeRobot~\cite{cadene2024lerobot}. We compare with GR00T N1.6~\cite{bjorck2025gr00t}, X-VLA~\cite{zheng2025x}, and MolmoAct2~\cite{fang2026molmoact2}, training on the same demonstrations and using matched evaluation settings. For in-distribution \texttt{Tape-to-Clay}, we evaluate five episodes per tape color (15 in total). All other real-world evaluations use 10 episodes per setting.

\noindent\textbf{Task.}
For \texttt{Potato-to-Plate}, we collect 40 demonstrations of picking up a potato and placing it on the middle of three circular plates. Zero-shot evaluation tests (1) placing the potato on a square plate in the middle, (2) placing it on the rightmost square plate, and (3) placing a lemon on a square plate. For \texttt{Tape-to-Clay}, we collect 45 demonstrations, 15 per tape color, of placing a red, blue, or yellow tape roll on white clay. The tapes are always arranged in red--blue--yellow order during training. With the tapes reordered to blue--red--yellow, we evaluate (1) blue tape on white clay, (2) yellow tape on white clay, and (3) blue tape on an unseen brown target. We also evaluate (4) yellow tape on a ball in yellow--red--blue order. All zero-shot variants are evaluated without additional training.

\noindent\textbf{Results.}
As shown in \figref{fig:exp_real_world}, RAIN achieves the highest OOD average in both task families. In \texttt{Potato-to-Plate}, all policies learn to place a potato on the middle plate. GR00T N1.6 and MolmoAct2 handle a change in plate shape, but their success drops when the target moves away from the middle. RAIN remains effective under this target-location shift. In \texttt{Tape-to-Clay}, RAIN also generalizes better to reordered source objects and changed targets. These results are consistent with the simulation results, where region-aware interactions transfer to new scene configurations.
\section{Conclusion}

We presented \textbf{Region-Aware Interaction Networks (RAIN)}, an action network that learns region-conditioned interactions for robotic task generalization. RAIN represents a task as a sequence of action types and target-region masks, which can be specified by a human or obtained through VLM-based planning and target-region tracking. TCE builds target-aware multi-view representations, PlanDiT predicts target-directed trajectories and actions, and the Transition Head determines when to advance to the next subtask. We further introduced \textbf{Reference-based Retargeting Strategy (RRS)}, which synthesizes smooth approaches to reference interactions with alternative target regions, diversifying training trajectories without collecting additional data. Experiments on LIBERO-Analogy show improved generalization across Adapt, Compose, and Decompose, alongside competitive performance on the original LIBERO suites. Real-world tabletop experiments also evaluate generalization to changes in objects, target locations, and scene layouts.
\section*{Acknowledgements}

\begin{sloppypar}
This work was supported by LG AI Research, 
the LG AI STAR Talent Development Program for Leading Large-Scale Generative AI Models in the Physical AI Domain (RS-2025-25442149), and the ``Advanced GPU Utilization Support Program'' funded by the Government of the Republic of Korea (Ministry of Science and ICT) (02-26-01-0057). 
It was additionally partly supported by grants from the IITP (RS-2021-II211343-GSAI/10\%, RS-2022-II220951-LBA/10\%, RS-2022-II220953-PICA/10\%, RS-2026-25553157-MIACC/10\%), the NRF (RS-2024-00353991-SPARC/15\%, RS-2023-00274280-HEI/15\%), the KEIT (RS-2025-25453780/15\%), and the KIAT (RS-2025-25460896/15\%), funded by the Korean government.
\end{sloppypar}

\bibliographystyle{IEEEtran}
\bibliography{main}

\clearpage
\useRomanappendicesfalse
\appendices
\section*{Appendix}
\pdfbookmark[0]{Appendix contents}{appendix-contents}
\suppressfloats[t]
\noindent\textbf{Contents}\par\smallskip
\begin{small}
\noindent
\begin{tabularx}{\linewidth}{@{}Xr@{}}
\hyperref[app:implementation_archive]{\ref*{app:implementation_archive}. Implementation and Supervision}
  & \pageref{app:implementation_archive} \\
\hyperref[app:perception_archive]{\ref*{app:perception_archive}. VLM Planning and Target Tracking}
  & \pageref{app:perception_archive} \\
\hyperref[app:benchmark_archive]{\ref*{app:benchmark_archive}. Simulation Benchmark and Metrics}
  & \pageref{app:benchmark_archive} \\
\hyperref[app:realworld_archive]{\ref*{app:realworld_archive}. Real-World Data and Evaluation}
  & \pageref{app:realworld_archive} \\
\end{tabularx}
\par
\end{small}

\section{Implementation and Supervision}
\label{app:implementation_archive}

\subsection{Observation and Action Interfaces}

In simulation, the policy receives third-person and wrist RGB images, target masks, an action type, and an eight-dimensional state. The state comprises the end-effector position, its axis-angle orientation, and the two gripper-finger positions. Each seven-dimensional action contains relative translation and rotation commands and a gripper command. The auxiliary plan contains eight three-dimensional waypoints; it is distinct from the action chunk sent to the controller.

The real-world platform uses three RGB views and a joint-space interface. Its recorded state contains six arm-joint positions and one gripper-joint value. Each action specifies six absolute arm-joint targets and a gripper command. The task interface, an action type paired with a target region, is shared across the two settings.

\subsection{Architecture and Optimization}

\begin{table}[!htb]
\centering
\caption{RAIN configuration. The visual and action-type encoders are frozen.}
\label{tab:archive_model_config}
\footnotesize
\setlength{\tabcolsep}{3pt}
\renewcommand{\arraystretch}{1.08}
\begin{tabularx}{\linewidth}{@{}lX@{}}
\toprule
Component & Setting \\
\midrule
Visual encoder & DINOv2-L/14; feature dimension 1024 \\
Action-type encoder & CLIP ViT-L/14 \\
Visual stages & Last three stages, $\ell\in\{2,3,4\}$ \\
TCE & TarLN and a cross-view Transformer block; parameters shared across stages \\
PlanDiT & $B=2$ cycles; six cross-attention and six self-attention layers \\
Prediction horizons & $H=16$ actions; $N=8$ plan waypoints \\
Consistency supervision & $K=4$ frames from the same subtask \\
Transition Head & Target-region average pooling and gated view fusion \\
Flow-matching steps & 4 \\
Executed actions per chunk & 8 \\
Subtask transition & Completion probability $>0.7$ for two consecutive predictions \\
\bottomrule
\end{tabularx}
\end{table}

Table~\ref{tab:archive_model_config} summarizes the model and inference settings. The three visual stages have the same spatial resolution and feature dimension. TarLN conditions each stage on its view's target mask, and TCE exchanges information between views before the stage features enter PlanDiT. Each decoder cycle consumes the three stage contexts in order. The action tokens can attend to the plan tokens, whereas the plan tokens cannot attend to the action tokens.

\begin{figure*}[!t]
\centering
\includegraphics[width=\textwidth]{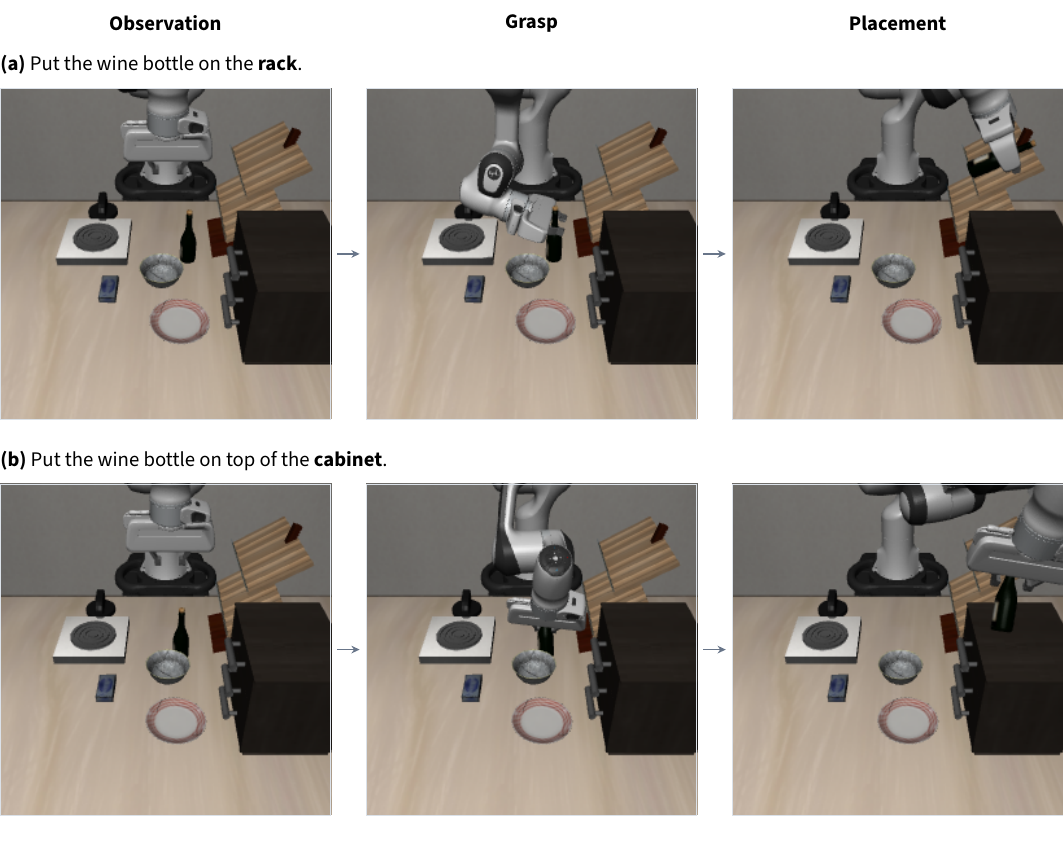}
\caption{\textbf{Different grasps for different destinations.} Successful LIBERO Goal demonstrations of placing the same wine bottle on a rack (top) or a cabinet (bottom), shown from the third-person view.}
\label{fig:destination_grasp}
\end{figure*}

Since the grasp can depend on the placement destination (Fig.~\ref{fig:destination_grasp}), we add a mask embedding for the target placement region.

The Transition Head uses the final-stage TCE features. Masked average pooling divides the sum of target-region features by the target area, so the pooled representation does not scale with object size. When a view has an empty target mask, the head uses global average pooling for that view. A learned gate combines the view features before concatenation with the CLIP action-type feature and completion prediction. The head is trained with weighted binary cross entropy.

\begin{table}[!htb]
\centering
\caption{Training settings for simulation and real-world experiments.}
\label{tab:archive_training}
\footnotesize
\setlength{\tabcolsep}{3pt}
\renewcommand{\arraystretch}{1.1}
\begin{tabularx}{\linewidth}{@{}Xcc@{}}
\toprule
Setting & Simulation & Real world \\
\midrule
RGB views & 2 & 3 \\
Input resolution & $224\times224$ & $448\times448$ \\
Total batch size & 1024 & 64 \\
Action-training iterations & 100k & 30k \\
\midrule
\multicolumn{3}{@{}l}{Shared optimization} \\
Optimizer & \multicolumn{2}{c}{AdamW; weight decay 0.01} \\
Learning rate & \multicolumn{2}{c}{$10^{-4}\rightarrow10^{-6}$, cosine decay} \\
Warmup & \multicolumn{2}{c}{2,000 iterations} \\
PlanDiT loss weights & \multicolumn{2}{c}{$\lambda_{\mathrm{fm}}=1$, $\lambda_{\mathrm{cons}}=0.1$} \\
\bottomrule
\end{tabularx}
\end{table}

Table~\ref{tab:archive_training} lists the training settings. Action training updates the region-conditioned encoder and PlanDiT while keeping DINOv2 and CLIP frozen. Transition Head training then freezes the action model and updates only the head. We reserve 10\% of the data for validation. The maximum budget is 30 epochs, approximately 7.5k iterations in simulation. Selected completed samples receive weight 1.5 and the remaining samples weight 1. Mask augmentation is applied during both action and Transition Head training.

\subsection{Subtask Supervision and Retargeting}

Task decomposition provides the ordered action types and targets. Recorded gripper motion determines the subtask intervals: closing, opening, and stabilization indicate grasp and release events, while changes or stops in translation and rotation delimit other interactions. The completion event determines the interval endpoint, which receives a positive completion label. Completion is conditioned on the target mask and action type, so the same observation can correspond to different completion labels for different interactions. To obtain simulation training masks, we replay recorded demonstrations and extract the simulator's segmentation at each frame. We exclude episodes that do not complete the task during replay. SAM~2 tracks the targets in the real-world demonstrations.

RRS constructs an alternative training target at a recorded observation. Candidate retargets must be compatible with the current gripper state; for example, placing an object requires a held object and a feasible destination. A smooth approach joins a recorded reference interaction at a tangent junction, after which the recorded reference actions are retained. The observation is paired with the alternative target mask and the synthesized action sequence. This changes the target-conditioned supervision without collecting another robot demonstration.
\begin{figure*}[t]
\centering
\includegraphics[width=\linewidth]{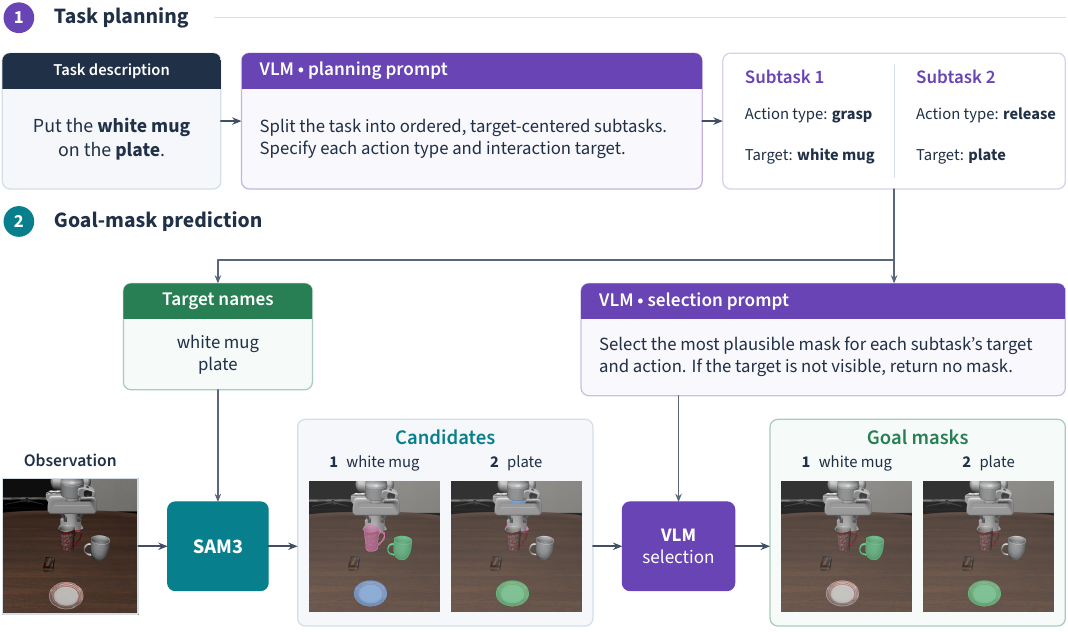}
\caption{\textbf{VLM-based goal-mask prediction.} The VLM decomposes the instruction into action--target pairs. SAM~3 proposes candidate masks using the observation and target names, and the same VLM selects the goal mask for each subtask. VLM prompts are abbreviated. For clarity, only the third-person observation is shown for this pick-and-place interaction; reference images and online tracking are omitted.}
\label{fig:archive_vlm_pipeline}
\end{figure*}

\section{VLM Planning and Target Tracking}
\label{app:perception_archive}

\subsection{From an Instruction to Targeted Interactions}

Figure~\ref{fig:archive_vlm_pipeline} illustrates how a task instruction becomes a sequence of target masks. A VLM first decomposes the instruction into ordered, target-centered subtasks. The planning prompt asks for the action type and interaction target of each subtask while preserving the instruction's spatial relations and execution order. References such as ``it'' are resolved to named objects. A pick-and-place instruction yields a grasp of the source object followed by a release at the destination.

The prompt includes examples of moving, pushing, opening, closing, and turning on objects. A shared conversion step expands a move into grasp and release primitives and formats the plan for RAIN. Planning is performed once per instruction.

\subsection{Candidate Generation and Selection}

Following AgentRVOS~\cite{jin2026agentrvos}, SAM~3 proposes masks from the current RGB images using target names. The selection prompt asks the same VLM to choose the candidate that best matches each subtask's target and action, using the scene and spatial relations to distinguish similar objects. The selected masks become the goal masks passed to RAIN. The views observe the same scene, but a target need not be visible in every view; the prompt permits an absent-target response instead of forcing a selection.

For simulation, a shared object catalog supplies object and manipulation-part names, aliases, and reference images. These inputs help distinguish unfamiliar objects and identify the part relevant to an action, such as a door rather than its enclosing appliance. Target names form the text queries to SAM~3; reference images support VLM candidate selection.

\subsection{Tracking and Closed-Loop Execution}

SAM~3 tracks the selected target masks online as new observations arrive. RAIN consumes these masks with RGB images, the action type, and the robot state to generate action chunks. Its Transition Head determines when to advance to the next subtask. The VLM is called for target acquisition or reacquisition, not for every tracking update.

When a target is absent or uncertain in one view, its mask is left empty and acquisition is retried on later observations. Cross-view checks maintain target identity between the third-person and wrist views. An empty mask alone does not trigger a subtask transition.
\section{Simulation Benchmark and Metrics}
\label{app:benchmark_archive}

\subsection{Training and Evaluation Splits}

RAIN is trained on the 40 tasks from LIBERO Spatial, Object, Goal, and Long using demonstrations that successfully complete the task during replay. Each suite contains ten tasks. LIBERO-Analogy evaluates transfer of these interactions to 60 tasks, divided equally among Adapt, Compose, and Decompose, without additional task-specific training. Table~\ref{tab:archive_coverage} lists the evaluation coverage.

\begin{table}[t]
\centering
\caption{Evaluation coverage. Every task uses 50 episodes.}
\label{tab:archive_coverage}
\footnotesize
\setlength{\tabcolsep}{4pt}
\begin{tabularx}{\linewidth}{@{}Xrr@{}}
\toprule
Benchmark / suite & Tasks & Episodes \\
\midrule
LIBERO Spatial & 10 & 500 \\
LIBERO Object & 10 & 500 \\
LIBERO Goal & 10 & 500 \\
LIBERO Long & 10 & 500 \\
\midrule
LIBERO-Analogy Adapt & 20 & 1,000 \\
LIBERO-Analogy Compose & 20 & 1,000 \\
LIBERO-Analogy Decompose & 20 & 1,000 \\
\bottomrule
\end{tabularx}
\end{table}

\noindent\textbf{Adapt.}
Tasks change the objective, object set, or scene layout while reusing learned interactions. The relevant object or destination can differ from the one associated with a familiar training scene, so following the original trajectory need not satisfy the new instruction.

\noindent\textbf{Compose.}
Tasks combine learned interactions into a longer sequence. Success requires completion of the requested compound task. Where an instruction specifies an order, the benchmark checks that order in addition to the final configuration.

\noindent\textbf{Decompose.}
Tasks request a subset of an interaction sequence learned during training. Completing the requested part is insufficient if the policy also performs a prohibited continuation. The benchmark therefore distinguishes the required outcome from outcomes associated with the omitted part of the original task.

\subsection{Success and Tracking Metrics}

Table~\ref{tab:vlm_sr_jf} reports success and mask quality from the same closed-loop episodes. Success rate is the fraction of episodes satisfying the task criterion. Region similarity $\mathcal{J}$ is the intersection over union of the predicted and GT masks, and contour accuracy $\mathcal{F}$ is the boundary F-measure. Tracking $\mathcal{J}\&\mathcal{F}$ is the mean of these two scores.

We pool valid visible-target observations across both views, target roles, and tasks, averaging region and contour scores over their respective valid observations. The overall $\mathcal{J}\&\mathcal{F}$ therefore weights suites by their contributing observations rather than taking an unweighted mean of suite scores.
\section{Real-World Data and Evaluation}
\label{app:realworld_archive}

\subsection{Robot, Cameras, and Recorded Data}

The ROBOTIS OMY-F3M has a six-degree-of-freedom arm and a two-finger gripper. An Intel RealSense D405 provides the wrist view, and two RealSense D435 cameras provide the third-person and top-down views. Demonstrations contain synchronized RGB observations, joint states, actions, and end-effector poses recorded through ROS and stored in LeRobot format~\cite{cadene2024lerobot}.

The robot client sends absolute joint targets at 30\,Hz. It obtains an action chunk from the policy server and executes a prefix before querying the policy again. The three RGB views are resized to $448\times448$ for RAIN, which is trained for 30k iterations with total batch size 64.

\subsection{Task Definitions}

\texttt{Potato-to-Plate} contains 40 demonstrations of placing a potato on the middle of three circular plates. The in-distribution evaluation retains this target relation. The three zero-shot settings change the target shape, target position, or manipulated object, as specified in Table~\ref{tab:archive_real_tasks}. Each setting uses ten evaluation episodes.

\texttt{Tape-to-Clay} contains 45 demonstrations, with 15 for each of the red, blue, and yellow tape rolls. The tapes are arranged in red--blue--yellow order during training, and the destination is white clay. In-distribution evaluation uses five episodes per color, for 15 episodes in total. Each of the four zero-shot settings uses ten episodes. No additional demonstrations from these zero-shot settings are used for training.

\begin{table*}[!t]
\centering
\caption{Real-world evaluation settings. Orders list the tape arrangement. OOD settings are evaluated without additional training.}
\label{tab:archive_real_tasks}
\footnotesize
\setlength{\tabcolsep}{4pt}
\renewcommand{\arraystretch}{1.18}
\begin{tabularx}{\textwidth}{@{}llXr@{}}
\toprule
Task family & Setting & Requested interaction and scene configuration & Episodes \\
\midrule
Potato-to-Plate & ID & Potato to the middle circular plate among three circular plates & 10 \\
 & OOD-1 & Potato to a square plate in the middle position & 10 \\
 & OOD-2 & Potato to the rightmost square plate & 10 \\
 & OOD-3 & Lemon to a square plate & 10 \\
\midrule
Tape-to-Clay & ID & Red, blue, or yellow tape to white clay; red--blue--yellow order & $3\times5$ \\
 & OOD-1 & Blue tape to white clay; blue--red--yellow order & 10 \\
 & OOD-2 & Yellow tape to white clay; blue--red--yellow order & 10 \\
 & OOD-3 & Blue tape to an unseen brown target; blue--red--yellow order & 10 \\
 & OOD-4 & Yellow tape to a ball; yellow--red--blue order & 10 \\
\bottomrule
\end{tabularx}
\end{table*}

\subsection{Comparison Protocol}

RAIN, GR00T N1.6, X-VLA, and MolmoAct2 use the same demonstrations within each task family and are evaluated on the same robot with matched camera and action interfaces. The baseline policies receive RGB observations, the instruction, and robot state. RAIN receives RGB observations, robot state, an action type, and target masks.

\begin{figure*}[!t]
\centering
\includegraphics[width=\textwidth]{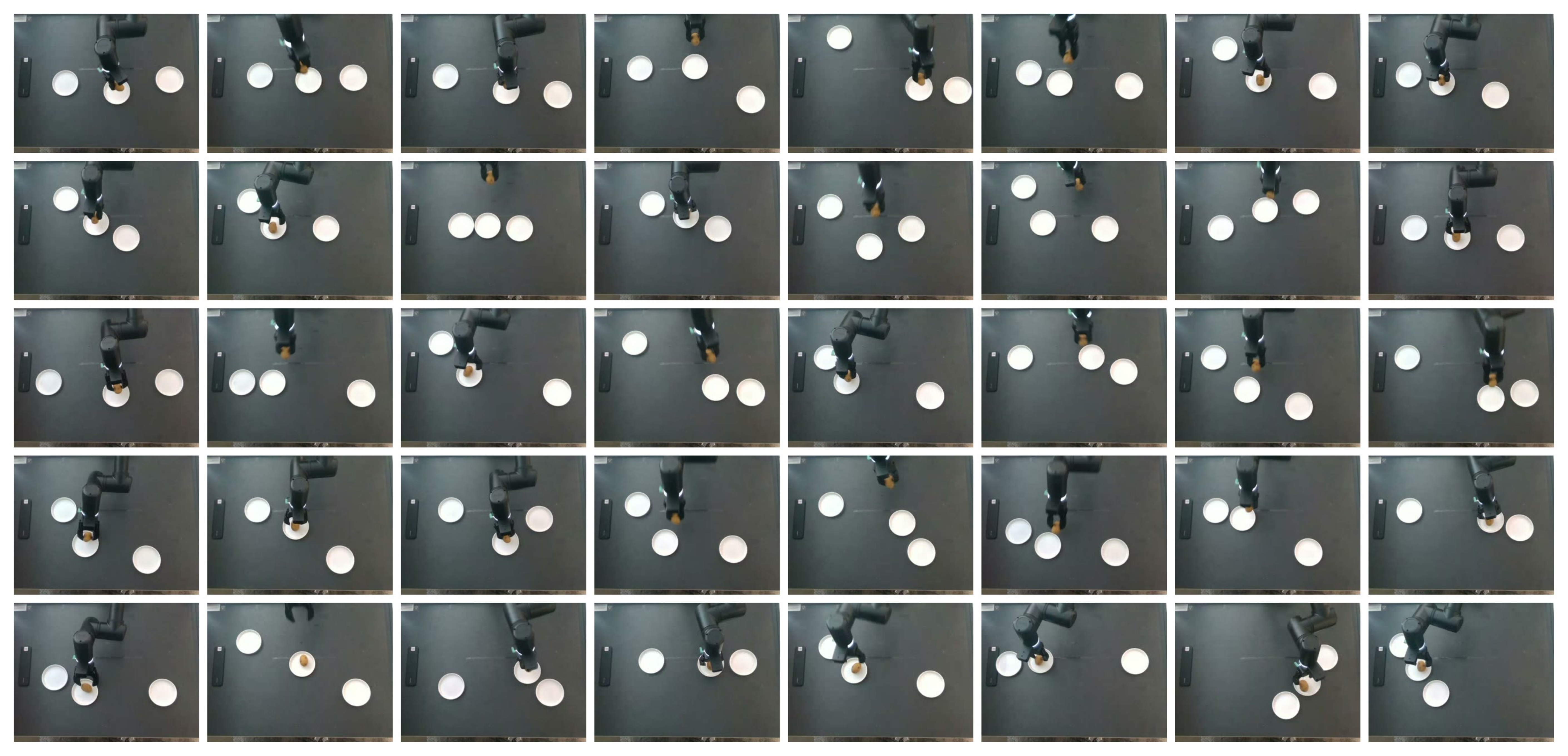}
\caption{\textbf{\texttt{Potato-to-Plate} training data.} One top-view frame is shown for each of the 40 demonstrations, ordered by episode index from left to right and then top to bottom.}
\label{fig:archive_potato_data}
\end{figure*}

\begin{figure*}[p]
\centering
\includegraphics[width=\textwidth,height=0.81\textheight,keepaspectratio]{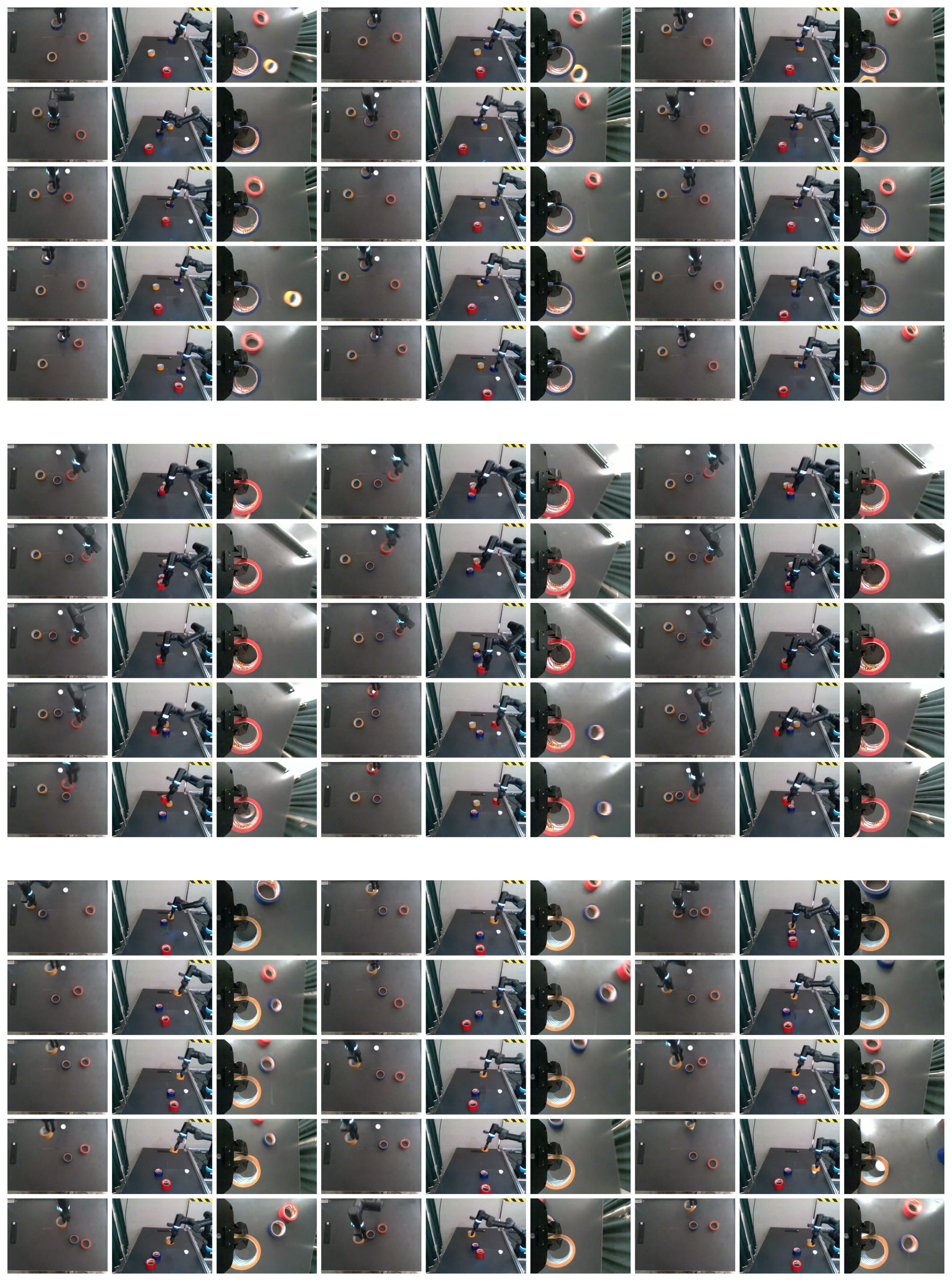}
\caption{\textbf{\texttt{Tape-to-Clay} training data.} The three bands show blue, red, and yellow tape variants, with 15 demonstrations per variant. Within each episode, the views are ordered top-down, third-person, and wrist.}
\label{fig:archive_tape_data}
\end{figure*}

Figure~\ref{fig:exp_real_world} presents the robot setup and quantitative evaluation results. Figures~\ref{fig:archive_potato_data} and~\ref{fig:archive_tape_data} show the training data for the two task families, including the objects, layouts, and camera views in the demonstrations.

\end{document}